\documentclass[10pt,twocolumn,letterpaper]{article}

\usepackage{cvpr}              % To produce the CAMERA-READY version
\usepackage{graphicx}
\usepackage{amsmath}
\usepackage{amssymb}
\usepackage{booktabs}
\makeatletter
\@namedef{ver@everyshi.sty}{}
\makeatother
\usepackage{tikz}

\usepackage[pagebackref,breaklinks,colorlinks]{hyperref}
 \usepackage{multirow}

\usepackage[capitalize]{cleveref}
\crefname{section}{Sec.}{Secs.}
\Crefname{section}{Section}{Sections}
\Crefname{table}{Table}{Tables}
\crefname{table}{Tab.}{Tabs.}

\def\cvprPaperID{11508} % *** Enter the CVPR Paper ID here
\def\confName{CVPR}
\def\confYear{2022}

\begin{document}

\title{An Aligned Multi-Temporal Multi-Resolution Satellite Image Dataset for Change Detection Research}

\author{Rahul Deshmukh\qquad Constantine J. Roros \qquad Amith Kashyap \qquad Avinash C. Kak \\
\large Purdue University \vspace{-.2em}\\
\normalsize
\{deshmuk5,~croros,~kashyap9,~kak\}@purdue.edu
}

% \and
\maketitle

%%%%%%%%% ABSTRACT
\begin{abstract}
This paper presents an aligned multi-temporal and
multi-resolution satellite image dataset for research in
change detection.  We expect our dataset to be useful to
researchers who want to fuse information from multiple
satellites for detecting changes on the surface of the earth
that may not be fully visible in any single satellite.  The
dataset we present was created by augmenting the SpaceNet-7
dataset \cite{van2021multi} with temporally parallel stacks
of Landsat and Sentinel images. The SpaceNet-7 dataset
consists of time-sequenced Planet  images recorded over
101 AOIs (Areas-of-Interest). In our dataset, for each of the
60 AOIs that are meant for training, we augment the Planet
 datacube with temporally parallel datacubes of Landsat
and Sentinel images.  The temporal alignments between the
high-res Planet  images, on the one hand, and the
Landsat and Sentinel images, on the other, are approximate
since the temporal resolution for the Planet images is one
month --- each image being a mosaic of the best data
collected over a month.  Whenever we have a choice regarding
which Landsat and Sentinel images to pair up with the Planet
 images, we have chosen those that had the least cloud
cover.  A particularly important feature of our dataset is
that the high-res and the low-res images are spatially
aligned together with our MuRA framework presented in this
paper.  Foundational to the alignment calculation is the
modeling of inter-satellite misalignment errors with
polynomials as in NASA's AROP \cite{arop2009} algorithm.  We
have named our dataset MuRA-T for the MuRA framework that is
used for aligning the cross-satellite images and "T" for the
temporal dimension in the dataset.
   
\end{abstract}

\section{Introduction}
\label{sec:intro}

Satellites are a rich source of data for identifying and
tracking significant changes on the surface of the earth.
Such changes are of great concern to a large variety of
people and that includes scientists, urban planners, those
engaged in damage assessment and mitigation planning when
natural disasters strike, etc.

For obvious reasons, detecting change requires at least two
images, before and after the event that is believed to have
caused the change.  More generally, if the goal is to
understanding the temporal evolution of the change taking
place, one would want a sequence of images in the form of a
time series for the same geographical region on the ground.

Such time series sequences of images can now be constructed
at medium resolution (around 4m) with the data made
available by Planet. As demonstrated by the SpaceNet-7
dataset, this resolution is adequate for identifying and
tracking changes related to buildings in urban areas.  For
each geographic area, the SpaceNet-7 dataset contains Planet
images, one per month (that may not always be consecutive),
over a span of around 24 months.

While SpaceNet-7 is a huge step forward in what's needed for
developing algorithms that can track changes at the
granularity of individual buildings in an urban area, our
own interest includes the geographic context of the urban
areas that contain the buildings.  That is, we are as
interested in the changes taking place in the landmasses
surrounding the buildings as we are in the purely urban
scenes consisting of just the buildings.

To that end, our dataset augments the SpaceNet-7 dataset
with temporally parallel Landsat and Sentinel datacubes of
images.  That is, for each Planet image in a temporal stack,
our dataset provides a Landsat image and a Sentinel image at
{\em roughly the same time}.  Each Planet image in the stack
represents the best of the Planet data captures over a
period of one month.  In other words, a Planet image is not
an instantaneous snapshot of what is on the ground, but a
mosaic of the best of what was seen at each point on the
ground during a span of one month.  In keeping with this
spirit, when we choose a low-res image, Landsat or Sentinel,
to go along with a Planet image, we try to choose the best
of the low-res images during that month on the basis of
primarily the extent of the cloud cover.\footnote{Our
  dataset includes temporally and spatially aligned
  Landsat-8 and Sentinel-2 images for all 60 SpaceNet-7 AOIs that are meant
  for training.}
  % The dataset also includes temporally and  spatially aligned Sentinel images for the 10 AOIs for which   those images were available/downloadable at the time we   last updated our dataset.}

What makes our dataset special is the fact that the low-res
images, Landsat and Sentinel, are spatially aligned with the
Planet images.  Since the images from the different
satellites are at different resolutions, one of the first
steps in the alignment process is deciding what {\em working
  resolution} to use for all the images for the purpose of
data alignment.

For aligning images from any pair of disparate satellites,
we have three choices: (1) Downsample the high-res images so
that its new sampling rate corresponds to that of the
low-res images; (2) Use super-resolution or image-to-image
translation neural networks to upsample the low-res images
so that its sampling rate corresponds to that of the
high-res images; and (3) Use a combination of downsampling
and upsampling for all the images from the different
satellites to correspond to some common agreed-upon sampling
rate.  As to which of these strategies to use depends on the
scale at which one would like to detect temporal changes on
the ground.  For the version of the dataset we are providing
at the moment, the alignment is carried out by downsampling
the Planet images to correspond to the sampling rates of the
Landsat and Sentinel data.

Our basic logic for cross-satellite image alignment was
inspired by NASA's AROP \cite{arop2009} algorithm in which the misalignment
errors are modeled by a polynomial. The degree of this
polynomial can be adapted to the required alignment
precision.  In the current dataset, the alignment accuracy
was measured by the reprojection error for the largest
possible inlier set of the corresponding tie-points between
the two images being aligned with each other.  This is a
standard approach to measuring the misalignment error
between a pair of images on a relative basis --- meaning
just with respect to each other as opposed to also with
respect to the features on the ground.  In future versions
of our dataset, our performance numbers related to alignment
precision will be augmented with those that describe
absolute accuracy with respect to the ground control points
(GCP) in those geographic regions where such land references
are available.\footnote{For large-scale image alignment
  experiments, it must be relatively easy to identify the
  GCPs in the satellite images.  For several regions in the
  US and other parts of the world, GCPs are made available
  by USGS for Landsat images.  The reason we have not yet
  incorporated these in the measurement of absolute
  alignment precision is because those GCPs come with
  neighbored depiction for the old 30m resolution Landsat-7
  data. It is an open research question at the moment
  whether we can super-resolve those neighborhood patterns
  to, say, the 15m resolution data for Landsat-8 images.}
As reported in \cref{sec:alignment_results}, all our
reprojection-error based alignment accuracies are at the
sub-pixel level.

The cross-satellite image alignment process as described
above was carried out with an algorithmic framework that,
for convenience, we have named MuRA for ``Multi-Resolution
Alignment''.  Since MuRA has played a central role in the
creation of our dataset, we have named it the MuRA-T Dataset
in which ``T'' refers to the temporal dimension of the data.
To the best of what we know, MuRA-T is the first temporally
and spatially aligned multi-satellite dataset that can be
used for change detection research.

In the rest of the paper, \cref{sec:lit_survey} presents a
brief literature survey of the different datasets currently
available for satellite images and the computational models
that can be used for aligning the same-satellite and
cross-satellite images.  Since our dataset is an
augmentation of the SpaceNet-7 dataset, we devote
\cref{sec:spacenet7} to presenting the relevant highlights
of this well-known dataset and asking the reader to consult
the original SpaceNet-7 paper for the details.
\cref{sec:attributes} presents a summary of the more
significant attributes of the Landsat and Sentinel images
that are relevant to how we have created our dataset.
\cref{sec:datacube} presents the ``mechanics'' of how we
accessed the different repositories on the internet for
constructing the MuRA-T dataset.  \cref{sec:mura} present
the MuRA logic that was used for spatially aligning the
images in MuRA-T.  The alignment results are presented in
\cref{sec:alignment_results}.

\section{Literature Survey} \label{sec:lit_survey}

This section first briefly reviews several other publicly
available datasets of satellite images while, at the same
time, pointing out how MuRA-T differs from them.  Following
that, we present a brief review of the image alignment
algorithms for satellite images.

\subsection{Satellite Image Datasets} 

The satellite image datasets that have received the most
attention during the last few years are those related to the
various SpaceNet challenges.  The paper by Van Etten et
al. \cite{spacenet_orig} provides a nice review of the
datasets created for the first three challenges and the
evaluation metrics used.  In these datasets, consisting of
time-static imagery, the focus was primarily on either just
building extraction or the extraction of both buildings and
roads from high-res imagery.  WorldView images proved ideal
for these datasets because of the high ground resolution
(30cm and 50cm) and the high quality of the images. Creating
the ground truth labels for the buildings (and roads)
required semi-automated processing of the data.

The three datasets used in the DeepGlobe challenge
\cite{deepglobe2018} were meant for three different tasks:
building detection, road extraction, and landcover
classification.  Of the datasets provided, the one for
building detection is actually from the SpaceNet collection.
For road extraction and landcover classification, the
datasets provided were based on DigitalGlobe+Vivid imagery.

There is also the multi-view satellite dataset presented in
\cite{sn4} for the development of algorithms that can work
directly with off-nadir satellite images.  The data is based
on WorldView-2 images at 50cm resolution.  The off-nadir
look angles in this data range from directly overhead to
$54^\circ$ elevation.  This static-time dataset was again
meant for solving the building detection problem, but in
off-nadir imagery.

There is also the SpaceNet-6 satellite image dataset
designed for developing and testing object detection
algorithms under all-weather conditions \cite{sn6}.  What
makes this dataset distinctive is that it includes both
optical and SAR (Synthetic Aperture Radar) imagery. An
interesting insight gained from this dataset was that the
object detection algorithms that are pre-trained with
optical images and then further trained on SAR images
outperform the algorithms that have only the SAR data to
work with.

Another interesting satellite image dataset is the xBD
dataset \cite{gupta2019xbd} meant specifically for developing
algorithms for assessing damage to buildings.  The dataset
consists of before-and-after images from 19 natural disaster
sites and was sourced from Maxar's Open Data program.  There
is also the EarthNet 2021 dataset
\cite{requena2020earthnet2021} that contains Sentinel-2
imagery over a span of several months for predicting
localized climate changes. Although temporal, this dataset
does not provide multi-satellite data like what we do.

That brings us to the SpaceNet-7 \cite{van2021multi} dataset
that contains 101 AOIs, 60 for training and the rest for
testing and validation, of Planet imagery over roughly a
24-month period. On account of the importance of this
dataset to our own, we have devoted a separate section to
it, \cref{sec:spacenet7}.

%alignment
\subsection{Image Alignment}

When dealing with multi-view images, the algorithms that
have proved most effective for aligning a mix of images over
the same geographic area are based on the RPC (Rational
Polynomial Coefficient) model for the satellite cameras and
the logic of bundle-adjustment \cite{triggs1999bundle}
\cite{lourakis2009sba} for the actual alignment.  Although
the RPC model has 78 polynomial coefficients as parameters,
Grodecki \& Dial \cite{grodecki2003block} have demonstrated
that the misalignment error in satellite images can be
corrected by estimating only the bias correction terms in
the RPC model.

When the images to be aligned involve only the nadir views
of the earth --- as is the case with the Planet images and
the low-res Landsat and Sentinel images --- the misalignment
errors are more easily modeled as affine transformations or,
equivalently, as polynomial warps as in NASA's
AROP\cite{arop2009} algorithm.  We use the same approach for
modeling the misalignment errors in our cross-satellite
image alignment work presented in this paper.

%-------------------------------------------------------------------------
\section{The SpaceNet-7 Dataset} \label{sec:spacenet7}

As mentioned previously, the dataset we present in this
paper, MuRA-T, is an augmentation of the SpaceNet-7 dataset.
Therefore, a few comments about the SpaceNet-7 dataset are
in order.

\begin{figure}
    \centering
    \includegraphics[scale=0.2]{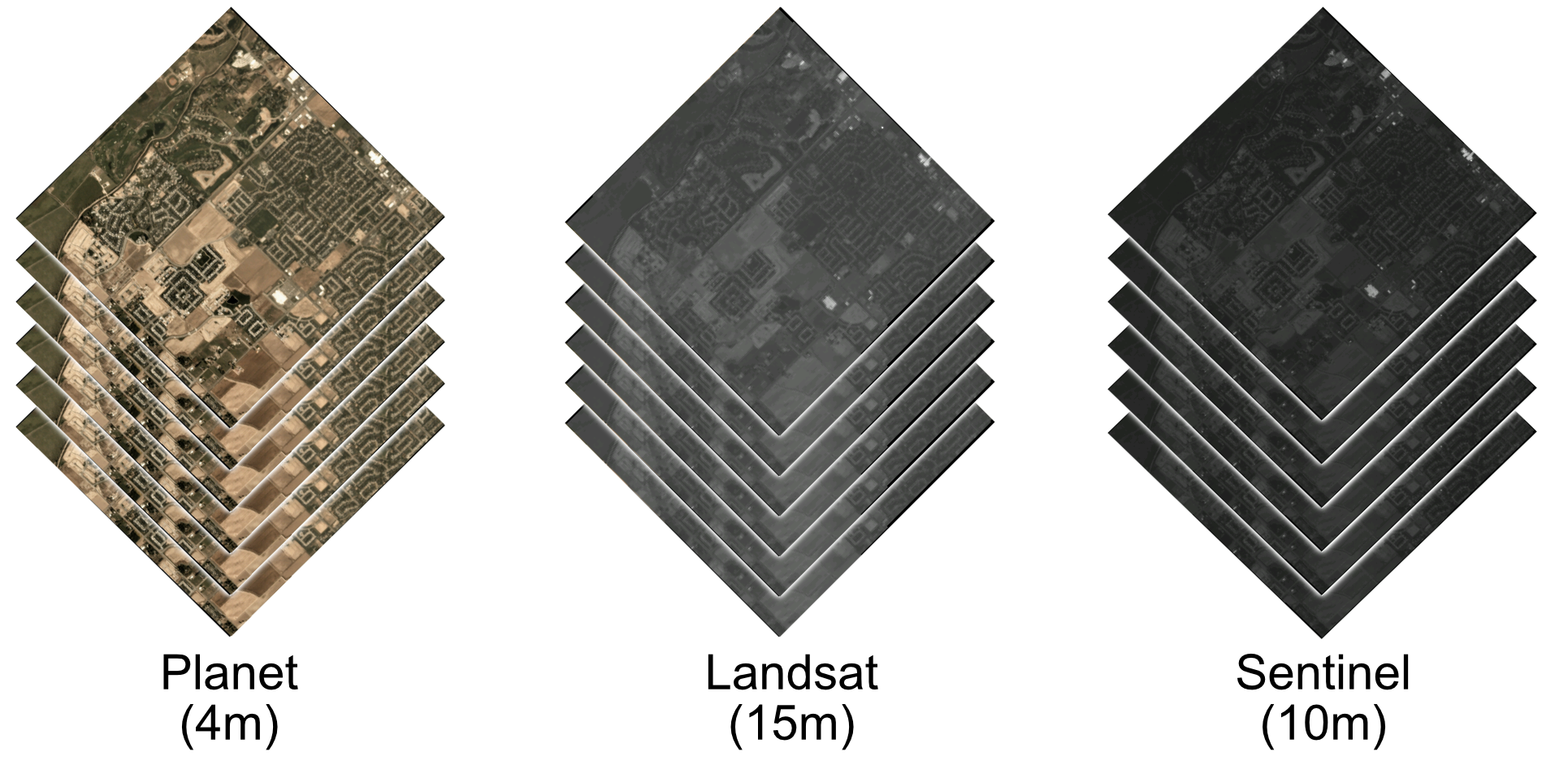}
    \caption{Three temporally paralleled and spatially
      aligned datacubes. The left datacube represents a
      stack of SpaceNet-7 Planet images, the center a stack
      of Landsat-8 images, and the right a stack of
      Sentinel-2 images.}

    \label{fig:murat_stacks}
\end{figure}

SpaceNet-7 consists of publicly available sequences of
Planet images, with each image representing the best of the
satellite captures over one month for the geographic area in
question.\footnote{The months may not be consecutive in a
  SpaceNet-7 stack over an AOI, but each image represents
  the best of satellite captures over a month.} This creates
a temporal stack of images as shown in \cref{fig:murat_stacks}, also referred to as a {\em
  datacube}, with a time resolution of one month (nominally
speaking).  SpaceNet-7 provides such datacubes for 101 AOIs
around the globe.  Of these, the data over 60 AOIs is meant
to be used for training, and the rest for testing and
validation.  On the average, each temporal stack spans
roughly 24 months.  The images in each stack are on-nadir
and are based on the data in the RGB bands with a spatial
resolution of roughly 4m on the ground. At this time, our dataset, MuRA-T, is based on just the 60
AOIs that are meant for training.

% We plan to increase this
% number to all 101 AOIs after we have received some feedback
% from the research community about current state of the
% dataset.

Since SpaceNet-7 is an ``urban development'' dataset, one of
its important goals is to promote research in building
detection algorithms. To that end, SpaceNet-7 provides
high-quality building annotations in all the images in the
dataset, with each building getting a separate label that
can be tracking in time.  The dataset comes with a footprint
mask for each building.  Each image also comes with an
``unusable data mask'' to designate the pixel blobs that are
not trusted on account of either the remaining cloud cover
in the images or because of unacceptable geo-reference
errors.  There are between 10,000 and 20,000 building
annotations in each image.

%-------------------------------------------------------------------------
\section{Some Relevant Attributes of the Landsat and Sentinel Imagery}  \label{sec:attributes}

As mentioned in the Introduction, our own research interests
go beyond just building detection --- we are also interested
in the geographic context of the buildings. That is, we are
as interested in the changes taking place in the landmasses
surrounding the buildings as in the purely urban scenes
consisting of just the buildings.

That's our main reason for augmenting the temporal stacks of
the Planet images in SpaceNet-7 with temporally parallel
stacks of Landsat and Sentinel images.  Since each Planet
image represents the best of the satellite captures over one
month, {\em it makes sense to pair each Planet image with
  the best possible Landsat image and the best possible
  Sentinel image recorded during the same month}.  By ``best
possible'', we mean the Landsat and Sentinel images with the
least cloud cover.  That's exactly what we have done in our
MuRA-T dataset. We used the F-MASK algorithm \cite{zhu2015}
for cloud detection.

Our augmentation of SpaceNet-7 with Landsat and Sentinel
images constitutes --- to the best of our knowledge --- a
first attempt at the creation of a temporally and spatially
aligned dataset of images from multiple satellites.  For
obvious reasons, this is going to lead to new opportunities
in research, in the sense that it would allow researches to
formulate more complex problems in change detection whose
solutions require data from multiple satellites
simultaneously.  Any algorithm development along these lines
would need to address the challenges created by the fact
that the images from the different satellites are likely to
sample the points on the ground at very different spatial
resolutions.

What makes the issue of having to deal with different
ground-sampling resolutions in different satellites even
more interesting is the fact there is no single value for
this spatial resolution for any given earth observation
satellite.  The data produced by such satellites is always
multi-spectral and each spectral band is characterized by
its own ground-sampling resolution value. \cref{table:landsat_band_details,table:sentinel_band_details} are a listing of the
different spectral bands imaged by the Landsat and Sentinel
satellites and their per pixel ground sampling resolution
values.  

Comparing the data presented in these two tables, the
highest ground sampling resolution value for Landsat-8 --
15m -- is for the Panchromatic (``Pan'' for short) band and
the same for Sentinel-2 is for the RGB bands, at 10m.

\begin{table}[!h]
    \centering
\resizebox{0.48\textwidth}{!}{\begin{tabular}{|c|l|c|}
    \hline
   Band & Band Description & Band wavelength \\
   No&&range ($\mu$m)\\
   \hline
   1 & 30 m Coastal/Aerosol  & 0.435-0.451\\
   \hline
   2& 30 m Blue& 0.452-0.512 \\
   \hline
   3& 30 m Green & 0.533-0.590\\
   \hline
   4& 30 m Red  & 0.636-0.673\\
   \hline
   5& 30 m Near-Infrared (NIR) & 0.851-0.879\\
   \hline
   6& 30 m Short Wavelength Infrared (SWIR)  & 1.566-1.651\\
   \hline
   7 & 30 m SWIR 2  & 2.107-2.294\\
   \hline
   8& 15 m Panchromatic  & 0.503-0.676\\
   \hline
   9& 30 m Cirrus & 1.363-1.384\\
   \hline
   10& 100 m Thermal Infrared Sensor (TIRS) 1& 10.60-11.19\\
   \hline
   11& 100 m TIRS 2 & 11.50-12.51\\
   \hline
   \end{tabular}}
    \vspace{-8pt}
    \caption{Landsat-8 band details}
    \label{table:landsat_band_details}
\end{table}

\begin{table}[!h]
    \centering
\resizebox{0.48\textwidth}{!}{\begin{tabular}{|c|l|c|c|}
    \hline
   Band & Band Description & Central & Bandwidth\\
   No &  & Wavelength & (nm)\\
   &&(nm) &\\
   \hline
   1 & 60 m Ultra blue  & 442.7& 21\\
   \hline
   2& 10 m Blue& 492.4& 66\\
   \hline
   3& 10 m Green & 559.8 & 36\\
   \hline
   4& 10 m Red  & 664.6& 31\\
   \hline
   5& 20 m Visible and Near-Infrared (VNIR) & 704.1& 15\\
   \hline
   6& 20 m VNIR  & 740.5& 15\\
   \hline
   7 & 20 m VNIR & 782.8& 20\\
   \hline
   8& 10 m VNIR  & 832.8& 106\\
   \hline
   8a& 20 m VNIR & 864.7& 21\\
   \hline
   9& 60 m Short Wavelength Infrared (SWIR) & 945.1& 20\\
   \hline
   10& 60 m SWIR & 1373.5& 31\\
   \hline
   11& 20 m SWIR & 1613.7& 91\\
   \hline
   12& 20 m SWIR & 2202.4& 175\\
   \hline
   \end{tabular}}
    \vspace{-8pt}
    \caption{Sentinel-2 band details}
    \label{table:sentinel_band_details}
\end{table}

%-------------------------------------------------------------------------
\section{Creating Temporally Paralleled Landsat And Sentinel Datacubes}\label{sec:datacube}

As one might expect, the key to temporally pairing
low-resolution data with the Planet images in the SpaceNet-7
stacks is finding Landsat and Sentinel images at the same
location and at times that correspond to the month
designators for the Planet images.  As it turns out, the
Lat/Long coordinates of each AOI are sufficient for finding
the corresponding Landsat images from the AWS Registry of
Open Data for these images \cite{landsat8aws}.

Images in the AWS Landsat registry are cataloged by their
WRS2 path and rows index values.\footnote{WRS2 is a
  Worldwide Reference System. It is analogous to UTM
  Zones. Basically WRS2 is a tiling system for the earth's
  surface in which the WGS84 ellipsoidal model of the earth
  is broken into several tiles indexed by their path and row
  coordinates. Landsat provides images in WRS2.}  Each image
is tagged with its acquisition date. Given the Lat/Long
coordinates associated with a Planet image, it is relatively
straightforward to find its corresponding WRS2 path and row
values.  We automated this process by writing a Python
script that compares the corner coordinates of each
SpaceNet-7 AOI with the WRS2 path and row shapefile provided
by NASA \cite{landsatwrs2shp}. Using the metadata list for
all Landsat scenes in the AWS registry, the URLs to the
relevant Landsat images can be found easily and, through
those URLs, the images can be downloaded with the AWS CLI.
In terms of their sizes, the SpaceNet-7 AOIs are much
smaller than the Landsat tiles. Therefore, in MuRA-T, the
downloaded Landsat images are clipped in order to correspond
to the AOIs.

Sentinel-2 images were accessed in a similar manner using the Google Cloud Platform with the help of BigQuery API.
Using these APIs, any Sentinel-2
imagery matching the SpaceNet-7 AOIs dates and locations can
be downloaded with the help of a custom python script.

\section{Spatially Aligning the Multi-Res Images}\label{sec:mura}

If the goal is to draw inferences from the data coming from
multiple satellites, each satellite characterized by its own
ground sampling distance, it is important for the
multi-satellite images at each time step to be aligned with
one another. Applications such as change detection can be
sensitive to even the slightest misalignments. The goal is
always to align the images with as high a sub-pixel
precision as possible.  To that end, we have developed a
framework of algorithms, named MuRA, for aligning together
multi-resolution images.  MuRA includes paths
for using super-resolution and image-to-image translation
networks for upsampling the low-res images to high-res.  The
different processing pathways in MuRA allow for the
following three different possibilities for cross-satellite
image alignment:

\begin{enumerate}
\item
In a mix of images from different satellites, we can
downsample all the images so that their ground sampling
resolution corresponds to that of the lowest-resolution
images.  For example, when aligning a 4m Planet image with a
15m Landsat image, we would first downsample the Planet
image to a 15m resolution.  We can think of this as the
least common denominator approach to dealing with the
alignment of multi-res images.

\item
There is a second possibility and that is a result of the
rapid advances that are currently being made in
super-resolution and image-to-image translation networks.
This possibility would call for upsampling the lower
resolution images to that of the highest-resolution image
and then applying the alignment logic to the data.  When
using super-resolution networks, the goal would be to make
fuzzy details look sharper --- in the same sense as in the
more traditional super-resolution algorithms that manipulate
the Fourier transform of an image by raising the transform
values near the limits of the Nyquist criterion.  On the
other hand, when using image-to-image translation networks,
one can imbue lower-resolution images with aspects of
structural sharpness as learned from the higher resolution
images. MuRA incorporates both types of these neural
networks.

\item
The third possibility is to use a combination of upsampling
and downsampling.  This approach would be appropriate for
situations when the resolution difference between the
highest-res and the lowest-res images in the mix is much too
large to handle with the either of the above two approaches.
\end{enumerate}

In the MuRA-T dataset being made available, we have only
used the processing path of MuRA for the first possibility
described above.  That is, for the version of the dataset we
are providing at the moment, the alignment is carried out by
downsampling the Planet images to correspond to the sampling
rates of the Landsat and Sentinel data.\footnote{We have
  also used MuRA to align Landsat and Sentinel images with
  the high-res WorldView images.  However, WorldView is not
  yet a part of the MuRA-T dataset.}

\subsection{The MuRA Pathways Used in the MuRA-T Dataset}\label{sec:mura_arch}

\begin{figure}[t]
  \centering
%   \fbox{\rule{0pt}{0.5in} \rule{0.9\linewidth}{0pt}}
  \includegraphics[width=0.8\linewidth]{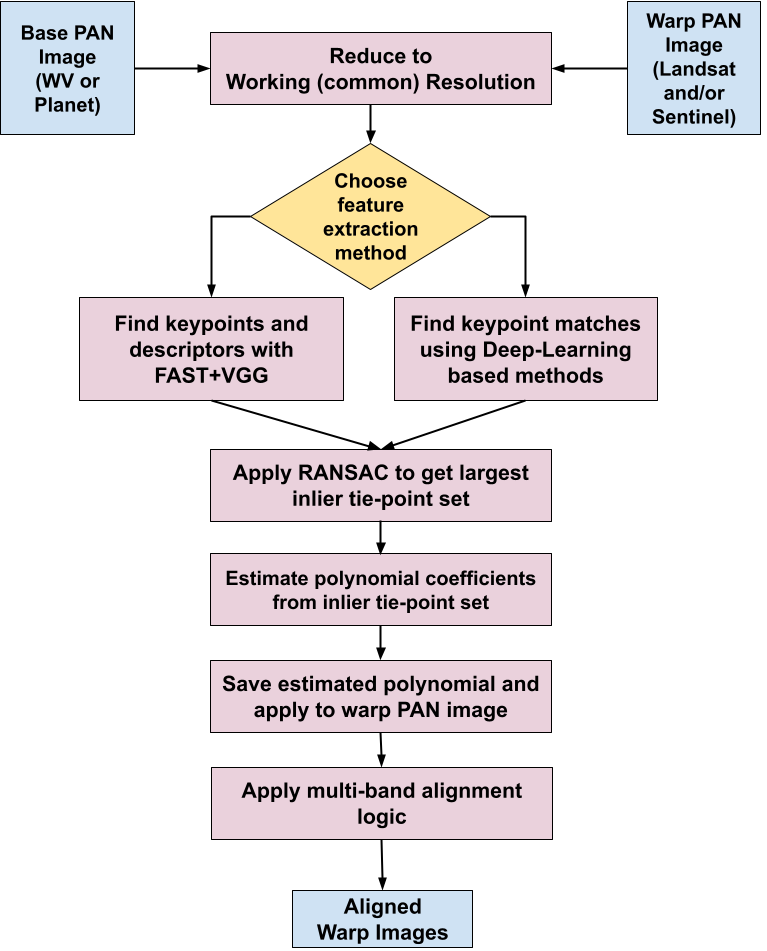}
   \caption{The computational pathway of the MuRA framework
     used for constructing the MuRA-T dataset.}
   \label{fig:mura}
\end{figure}

\cref{fig:mura} shows the section of the MuRA framework used
for constructing the MuRA-T dataset.  First, the higher
resolution image is downsampled so each image is at a common
working resolution (i.e. the lowest resolution among the
input images). Once re-sampled, we obtain tiepoints using
one of several possible feature extraction and matching
methods. Our framework is based on a plug-n-play design that
makes it easy for us to test different possible feature
extraction and the feature matching methods for image
alignment.  We used both FAST+VGG \cite{fast2006, vgg2015}
and SuperPoint+SuperGlue \cite{superpoint2017,
  superglue2019} algorithms for the alignments needed for
MuRA-T because these algorithms yielded the best
results. Further details are presented in
\cref{sec:pipelines}.

Once we obtain a set of tiepoints through feature matching,
we apply bundle adjustment \cite{triggs1999bundle} to this
set in order to estimate the coefficients of the polynomial
model used for the misalignment error.  An important
component of the bundle-adjustment logic is the RANSAC
algorithm \cite{fischler1981} for outlier rejection. RANSAC
is invoked iteratively in order to find the largest inlier
set of tiepoints for aligning a pair of images. The nature
of the polynomial model is described in
\cref{sec:polynomials}. Using the estimated polynomial
model, we re-sample the misaligned low-resolution image into
an aligned image.

\begin{figure*}[!h]
    \centering
    \includegraphics[width=\textwidth]{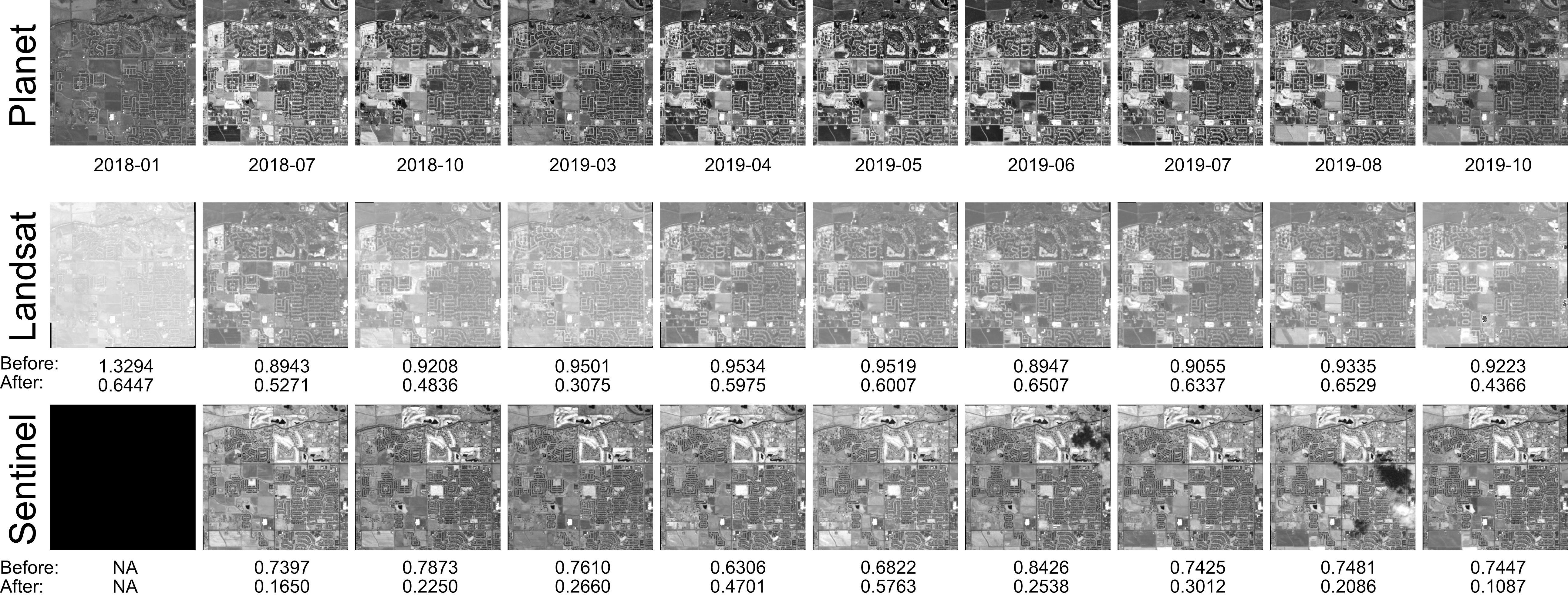}
    \caption{Shown in the top row are 10 of the 24 Planet
      images for the Meridian, Idaho, AOI. On account of the
      space constraints, we have shown only the first ten
      images in the Planet stack over the AOI. Note the
      Planet images in the SpaceNet-7 stacks are not always
      for consecutive months, as one can tell from
      month/year tags shown below the Planet images.  The
      second row shows the aligned Landsat images and the
      third row the aligned Sentinel images.  The Landsat
      and the Sentinel images were chosen as the
      best-of-the-month on the basis of containing the least
      cloud cover as detected by the F-MASK \cite{zhu2015}
      algorithm.  The before and after misalignment errors
      in terms of the RMSE values are also displayed for
      each of the images.}
    \label{fig:time_series}
\end{figure*}

%-------------------------------------------------------------------------
\subsection{Modeling the Misalignment Error with Polynomials} \label{sec:polynomials}

Modeling of the misalignment errors in the alignment logic
used for MuRA-T is based on NASA's AROP \cite{arop2009}
algorithm that was used to harmonize Landsat and Sentinel
images into a single unified dataset.  The polynomial
correction model can essentially be thought of as an affine
homography that is a relationship between the pixel
coordinates of the corresponding pixel positions in a pair
of images with the exception that the polynomial correction
model can use higher degree polynomials unlike affine
homographies.  Shown below are the three choices for such
polynomials. In the definitions shown, $(\hat{x}_w,
\hat{y}_w)$ represent the {\em warp image} (meaning the
misaligned image) pixel coordinates and $(x_b,y_b)$ the
pixel coordinates in the base image (meaning the reference
image).

\vspace{1ex}
\noindent Shift Correction Model:
\begin{equation}
\begin{split}
  \hat{x}_w &= a_1 + x_b\\
  \hat{y}_w &= b_1 + y_b
\end{split}
\label{eq:poly_shift}
\end{equation}
Affine Correction Model:
\begin{equation}
\begin{split}
  \hat{x}_w &= a_1 + a_2x_b + a_3y_b\\
  \hat{y}_w &= b_1 + b_2x_b + b_3y_b
\end{split}
\label{eq:poly_linear}
\end{equation}
%\vspace{0.1in}
Quadratic Polynomial Correction Model:
\begin{equation}
\begin{split}
  \hat{x}_w &= a_1 + a_2x_b + a_3y_b+ a_4x_b^2 + a_5x_by_b+ a_6y_b^2\\
  \hat{y}_w &= b_1 + b_2x_b + b_3y_b + b_4x_b^2 + b_5x_by_b + b_6y_b^2
\end{split}
\label{eq:poly_quad}  
\end{equation}

The goal of alignment is to estimate the coefficients
$\{a_i\}$ and $\{b_i\}$ given the corresponding tie points
in the two images.  This is accomplished by minimizing the
distance between the actual measurements for the coordinates
$(x_w, y_w)$ and their estimated values $(\hat{x}_w,
\hat{y}_w)$ as yielded by the alignment algorithm --- this
distance is frequently referred to as the {\em reprojection
  error}. Formally, the minimization problem is stated as:
%\cref{eq:argmin_reprojection_error}

\begin{equation}
\begin{split}
\{a_i^*\}, \{b_i^*\} = \operatorname*{argmin}_{\{a_i\},\{b_i\}} \Big{\{} \sum_{k=0}^{N-1} (x_w^{(k)} - \hat{x}_w^{(k)})^2 +\\ (y_w^{(k)} - \hat{y}_w^{(k)})^2\Big{\}}
\end{split}
\label{eq:argmin_reprojection_error}      
\end{equation}

\noindent where $k$ is an index for a pair of tie-point
correspondences among all the $N$ inliers used for computing
the reprojection error. Fortunately, the problem in
\cref{eq:argmin_reprojection_error} can be easily converted
to a linear least square problem as the polynomial
correction model is a linear function of the parameters
$\{a_i\}$ and $\{b_i\}$.

%-------------------------------------------------------------------------

\subsection{The Alignment Pipeline Used for MuRA-T} \label{sec:pipelines}

Since the new feature extraction and matching methods are
being reported constantly, as previously stated, the
computational steps in the image alignment pipeline shown in
\cref{fig:mura} are meant to operate on a plug-n-play basis.
This has allowed us to experiment with a large number of
possibilities for the feature extraction and feature
matching steps in image alignment.  The algorithms we tested
include: SIFT \cite{lowe1999object}, SURF
\cite{bay2008speeded}, ORB \cite{rublee2011orb}, STAR+BRIEF
\cite{agrawal2008star,calonder2010brief}, FAST for feature
extraction and VGG for matching, SuperPoint for feature
extraction and SuperGlue for matching, etc.  Of all these
methods, we obtained the best results with FAST+VGG and
SuperPoint+SuperGlue. These two approaches yielded the
largest number of tiepoints between a pair of images on the
average.  While SuperPoint+SuperGlue is an end-to-end deep
learning approach, FAST+VGG combines both traditional and
deep learning-based methods towards the goal of feature
extraction.

\subsection{Resampling and Multi-Band Alignment}
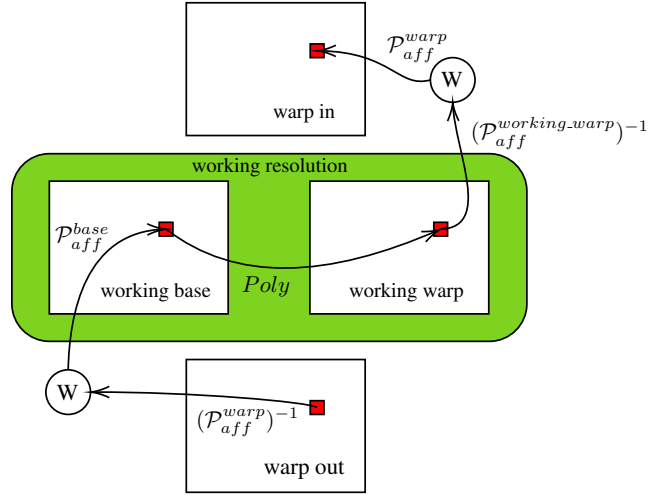
\begin{figure}[h]
\centering
\resizebox{0.5\textwidth}{!}{
\tikzset{every picture/.style={line width=0.75pt}} %set default line width to 0.75pt
\begin{tikzpicture}[x=0.75pt,y=0.75pt,yscale=-1,xscale=1]

%Rounded Rect [id:dp0952830759821901]
\draw  [fill={rgb, 255:red, 126; green, 211; blue, 33 }  ,fill opacity=1 ] (170.2,117.88) .. controls (170.2,105.8) and (180,96) .. (192.08,96) -- (448.32,96) .. controls (460.4,96) and (470.2,105.8) .. (470.2,117.88) -- (470.2,183.52) .. controls (470.2,195.6) and (460.4,205.4) .. (448.32,205.4) -- (192.08,205.4) .. controls (180,205.4) and (170.2,195.6) .. (170.2,183.52) -- cycle ;
%Shape: Rectangle [id:dp8056849907058883]
\draw  [fill={rgb, 255:red, 255; green, 255; blue, 255 }  ,fill opacity=1 ] (192,112) -- (296.2,112) -- (296.2,189.4) -- (192,189.4) -- cycle ;
%Shape: Rectangle [id:dp6011921343779998]
\draw  [fill={rgb, 255:red, 255; green, 255; blue, 255 }  ,fill opacity=1 ] (343.8,112) -- (448,112) -- (448,189.4) -- (343.8,189.4) -- cycle ;
%Shape: Rectangle [id:dp041119049820856235]
\draw  [fill={rgb, 255:red, 255; green, 255; blue, 255 }  ,fill opacity=1 ] (271.8,216) -- (376,216) -- (376,293.4) -- (271.8,293.4) -- cycle ;
%Shape: Rectangle [id:dp6571795937754166]
\draw  [fill={rgb, 255:red, 255; green, 255; blue, 255 }  ,fill opacity=1 ] (271.8,8) -- (376,8) -- (376,85.4) -- (271.8,85.4) -- cycle ;
%Shape: Rectangle [id:dp46110536693719584]
\draw  [fill={rgb, 255:red, 255; green, 0; blue, 0 }  ,fill opacity=1 ] (264,144) -- (256,144) -- (256,136) -- (264,136) -- cycle ;
%Shape: Rectangle [id:dp9405668351690153]
\draw  [fill={rgb, 255:red, 255; green, 0; blue, 0 }  ,fill opacity=1 ] (424,144) -- (416,144) -- (416,136) -- (424,136) -- cycle ;
%Shape: Rectangle [id:dp367502412616147]
\draw  [fill={rgb, 255:red, 255; green, 0; blue, 0 }  ,fill opacity=1 ] (352,248) -- (344,248) -- (344,240) -- (352,240) -- cycle ;
%Shape: Rectangle [id:dp3303325591293491]
\draw  [fill={rgb, 255:red, 255; green, 0; blue, 0 }  ,fill opacity=1 ] (352,40) -- (344,40) -- (344,32) -- (352,32) -- cycle ;
%Shape: Circle [id:dp7262720791136228]
\draw   (190,235) .. controls (190,227.82) and (195.82,222) .. (203,222) .. controls (210.18,222) and (216,227.82) .. (216,235) .. controls (216,242.18) and (210.18,248) .. (203,248) .. controls (195.82,248) and (190,242.18) .. (190,235) -- cycle ;
%Curve Lines [id:da04341100026671141]
\draw    (348,244) .. controls (343.71,240.7) and (274.41,234.79) .. (217.71,234.99) ;
\draw [shift={(216,235)}, rotate = 359.66] [color={rgb, 255:red, 0; green, 0; blue, 0 }  ][line width=0.75]    (10.93,-3.29) .. controls (6.95,-1.4) and (3.31,-0.3) .. (0,0) .. controls (3.31,0.3) and (6.95,1.4) .. (10.93,3.29)   ;
%Curve Lines [id:da122995846950821]
\draw    (203,222) .. controls (203.66,189) and (212.81,141.63) .. (258.6,140.03) ;
\draw [shift={(260,140)}, rotate = 179.19] [color={rgb, 255:red, 0; green, 0; blue, 0 }  ][line width=0.75]    (10.93,-3.29) .. controls (6.95,-1.4) and (3.31,-0.3) .. (0,0) .. controls (3.31,0.3) and (6.95,1.4) .. (10.93,3.29)   ;
%Curve Lines [id:da9268612753806988]
\draw    (260,140) .. controls (300.8,171.4) and (353.8,168.91) .. (419.02,140.43) ;
\draw [shift={(420,140)}, rotate = 156.25] [color={rgb, 255:red, 0; green, 0; blue, 0 }  ][line width=0.75]    (10.93,-3.29) .. controls (6.95,-1.4) and (3.31,-0.3) .. (0,0) .. controls (3.31,0.3) and (6.95,1.4) .. (10.93,3.29)   ;
%Curve Lines [id:da901862230554896]
\draw    (420,140) .. controls (450.38,138.91) and (428.58,94.71) .. (427.07,67.64) ;
\draw [shift={(427,66)}, rotate = 88.58] [color={rgb, 255:red, 0; green, 0; blue, 0 }  ][line width=0.75]    (10.93,-3.29) .. controls (6.95,-1.4) and (3.31,-0.3) .. (0,0) .. controls (3.31,0.3) and (6.95,1.4) .. (10.93,3.29)   ;
%Shape: Circle [id:dp3199698448013053]
\draw   (414,53) .. controls (414,45.82) and (419.82,40) .. (427,40) .. controls (434.18,40) and (440,45.82) .. (440,53) .. controls (440,60.18) and (434.18,66) .. (427,66) .. controls (419.82,66) and (414,60.18) .. (414,53) -- cycle ;
%Curve Lines [id:da35566394523732403]
\draw    (414,53) .. controls (395.29,58.14) and (391.77,36.88) .. (349.94,36.02) ;
\draw [shift={(348,36)}, rotate = 0.29] [color={rgb, 255:red, 0; green, 0; blue, 0 }  ][line width=0.75]    (10.93,-3.29) .. controls (6.95,-1.4) and (3.31,-0.3) .. (0,0) .. controls (3.31,0.3) and (6.95,1.4) .. (10.93,3.29)   ;

% Text Node
\draw (202.63,235) node   [align=left] {\begin{minipage}[lt]{18.19pt}\setlength\topsep{0pt}
\begin{center}
W
\end{center}

\end{minipage}};
% Text Node
\draw (426.63,53) node   [align=left] {\begin{minipage}[lt]{18.19pt}\setlength\topsep{0pt}
\begin{center}
W
\end{center}

\end{minipage}};
% Text Node
\draw (340.5,280.5) node   [align=left] {\begin{minipage}[lt]{48.51pt}\setlength\topsep{0pt}
\begin{center}
warp out
\end{center}

\end{minipage}};
% Text Node
\draw (254,176.5) node   [align=left] {\begin{minipage}[lt]{65.39pt}\setlength\topsep{0pt}
\begin{center}
{\small working base}
\end{center}

\end{minipage}};
% Text Node
\draw (400,176.5) node   [align=left] {\begin{minipage}[lt]{65.39pt}\setlength\topsep{0pt}
\begin{center}
{\small working warp}
\end{center}

\end{minipage}};
% Text Node
\draw (340.5,72.5) node   [align=left] {\begin{minipage}[lt]{65.39pt}\setlength\topsep{0pt}
\begin{center}
{\small warp in}
\end{center}

\end{minipage}};
% Text Node
\draw (338.67,104) node   [align=left] {\begin{minipage}[lt]{95.43pt}\setlength\topsep{0pt}
{\small working resolution}
\end{minipage}};
% Text Node
\draw (276,242.73) node [anchor=north west][inner sep=0.75pt]    {$(\mathcal{P}_{aff}^{warp})^{-1}$};
% Text Node
\draw (194,134.73) node [anchor=north west][inner sep=0.75pt]    {$\mathcal{P}_{aff}^{base}$};
% Text Node
\draw (436,73) node [anchor=north west][inner sep=0.75pt]    {$(\mathcal{P}_{aff}^{working\_warp})^{-1}$};
% Text Node
\draw (387.33,25) node [anchor=north west][inner sep=0.75pt]    {$\mathcal{P}_{aff}^{warp}$};
% Text Node
\draw (302.67,165.96) node [anchor=north west][inner sep=0.75pt]    {$Poly$};

\end{tikzpicture}
}
\caption{Resampling using affine correction model. We align images at a common working resolution (shown in green). The correction model is denoted by the transformation \textit{Poly} which relates the working base with the working warp. Affine projection ($\mathcal{P}_{aff}$) and back projection ($\mathcal{P}_{aff}^{-1}$) are used for transformations from pixel to world space (\textit{W} in circle) and vice-versa.}
\label{fig:aff_resampling}
\end{figure}

As discussed in \cref{sec:polynomials}, the correction model is a function of the base image coordinates at the working resolution.  After determining the best polynomial for the PAN-to-PAN alignment between the reference image and the low-res image, next we need to align all the other bands of the warp image in question.  To that end, each band of the warp image is aligned by modifying the PAN-to-PAN polynomial using the metadata for that band. This calculation requires us to go through several transformations as shown in \cref{fig:aff_resampling}. In this figure, “warp out” refers to the aligned version of the band image in question and “warp in” refers to the band image before alignment. Here is a listing of the computational steps: 
\begin{enumerate}

  \item For the pixel location in output warp image -- warp-out -- as shown in \cref{fig:aff_resampling} we use the affine back projection function ($(\mathcal{P}_{aff}^{warp})^{-1}$) supplied by the metadata to get the corresponding 2D world location.  

  \item Using the 2D world location from the previous step, we project it into the working base image using the affine projection function, supplied again by the metadata, for the working base image ($\mathcal{P}_{aff}^{base}$) and obtain the pixel coordinate $(x_b,y_b)$.  

  \item Using the pixel coordinates of the working base and the polynomial correction model explained in \cref{sec:polynomials}, we identify the corrected warp pixel coordinates $(\hat{x}_w,\hat{y}_w)$.

  \item Using the affine back projection function ($(\mathcal{P}_{aff}^{working\_warp})^{-1}$) of the working warp image, supplied by its metadata, we backproject the warp pixel coordinates $(\hat{x}_w,\hat{y}_w)$ to get their corresponding 2D world coordinates. 

  \item Finally, we project the 2D world coordinates from the previous step into the misaligned warp image, meaning the warp-in image, and obtain a pixel location. The value at this pixel location gets written into the output aligned warp image. 
\end{enumerate}

Using the above process we can produce resampled aligned images with any choice of interpolation method. 

%% figure for resampling
% resampling ends
%----------------------------------------

% \usepackage{multirow}
% \usepackage{multirow}
% \usepackage{multirow}
\begin{table*}[!h]
\centering
\resizebox{\textwidth}{!}{
\begin{tabular}{|c|c|cccc|cccc|}
\hline
\multirow{3}{*}{AOI} & \multirow{3}{*}{\begin{tabular}[c]{@{}c@{}}Low-Res Images \\ For Alignment\end{tabular}} & \multicolumn{4}{c|}{FAST+VGG} & \multicolumn{4}{c|}{SuperPoint+SuperGlue} \\ \cline{3-10} 
 &  & \multicolumn{2}{c|}{RMSE Before} & \multicolumn{2}{c|}{RMSE After} & \multicolumn{2}{c|}{RMSE Before} & \multicolumn{2}{c|}{RMSE After} \\ \cline{3-10} 
 &  & \multicolumn{1}{c|}{Avg} & \multicolumn{1}{c|}{Std Dev} & \multicolumn{1}{c|}{Avg} & Std Dev & \multicolumn{1}{c|}{Avg} & \multicolumn{1}{c|}{Std Dev} & \multicolumn{1}{c|}{Avg} & Std Dev \\ \hline
\multirow{2}{*}{\begin{tabular}[c]{@{}c@{}}Chula Vista, California, \\ United States\end{tabular}} & Landsat & \multicolumn{1}{c|}{0.801} & \multicolumn{1}{c|}{0.121} & \multicolumn{1}{c|}{0.197} & 0.158 & \multicolumn{1}{c|}{1.189} & \multicolumn{1}{c|}{0.171} & \multicolumn{1}{c|}{0.558} & 0.053 \\ \cline{2-10} 
 & Sentinel-2 & \multicolumn{1}{c|}{0.908} & \multicolumn{1}{c|}{0.211} & \multicolumn{1}{c|}{0.509} & 0.101 & \multicolumn{1}{c|}{1.336} & \multicolumn{1}{c|}{0.223} & \multicolumn{1}{c|}{0.625} & 0.03 \\ \hline
\multirow{2}{*}{\begin{tabular}[c]{@{}c@{}}Bentonville, Arkansas, \\ United States\end{tabular}} & Landsat & \multicolumn{1}{c|}{0.866} & \multicolumn{1}{c|}{0.161} & \multicolumn{1}{c|}{0.558} & 0.066 & \multicolumn{1}{c|}{1.102} & \multicolumn{1}{c|}{0.083} & \multicolumn{1}{c|}{0.639} & 0.031 \\ \cline{2-10} 
 & Sentinel-2 & \multicolumn{1}{c|}{0.971} & \multicolumn{1}{c|}{0.212} & \multicolumn{1}{c|}{0.529} & 0.085 & \multicolumn{1}{c|}{2.066} & \multicolumn{1}{c|}{2.692} & \multicolumn{1}{c|}{0.6} & 0.065 \\ \hline
\multirow{2}{*}{\begin{tabular}[c]{@{}c@{}}Atlanta, Georgia, \\ United States\end{tabular}} & Landsat & \multicolumn{1}{c|}{1.753} & \multicolumn{1}{c|}{0.393} & \multicolumn{1}{c|}{0.306} & 0.162 & \multicolumn{1}{c|}{2.032} & \multicolumn{1}{c|}{0.296} & \multicolumn{1}{c|}{0.598} & 0.025 \\ \cline{2-10} 
 & Sentinel-2 & \multicolumn{1}{c|}{1.004} & \multicolumn{1}{c|}{0.136} & \multicolumn{1}{c|}{0.46} & 0.079 & \multicolumn{1}{c|}{1.386} & \multicolumn{1}{c|}{0.128} & \multicolumn{1}{c|}{0.625} & 0.032 \\ \hline
\multirow{2}{*}{\begin{tabular}[c]{@{}c@{}}Rotterdam, Zuid-Holland, \\ Netherlands\end{tabular}} & Landsat & \multicolumn{1}{c|}{1.538} & \multicolumn{1}{c|}{0.353} & \multicolumn{1}{c|}{0.529} & 0.085 & \multicolumn{1}{c|}{1.737} & \multicolumn{1}{c|}{0.325} & \multicolumn{1}{c|}{0.61} & 0.048 \\ \cline{2-10} 
 & Sentinel-2 & \multicolumn{1}{c|}{1.172} & \multicolumn{1}{c|}{0.194} & \multicolumn{1}{c|}{0.522} & 0.073 & \multicolumn{1}{c|}{1.578} & \multicolumn{1}{c|}{0.164} & \multicolumn{1}{c|}{0.604} & 0.017 \\ \hline
\multirow{2}{*}{Dawm, Sudan} & Landsat & \multicolumn{1}{c|}{0.794} & \multicolumn{1}{c|}{0.121} & \multicolumn{1}{c|}{0.226} & 0.136 & \multicolumn{1}{c|}{1.31} & \multicolumn{1}{c|}{0.313} & \multicolumn{1}{c|}{0.547} & 0.051 \\ \cline{2-10} 
 & Sentinel-2 & \multicolumn{1}{c|}{1.02} & \multicolumn{1}{c|}{0.503} & \multicolumn{1}{c|}{0.503} & 0.075 & \multicolumn{1}{c|}{1.408} & \multicolumn{1}{c|}{0.154} & \multicolumn{1}{c|}{0.628} & 0.03 \\ \hline
\multirow{2}{*}{Mukono, Uganda} & Landsat & \multicolumn{1}{c|}{2.11} & \multicolumn{1}{c|}{0.221} & \multicolumn{1}{c|}{0.45} & 0.13 & \multicolumn{1}{c|}{2.323} & \multicolumn{1}{c|}{0.462} & \multicolumn{1}{c|}{0.619} & 0.027 \\ \cline{2-10} 
 & Sentinel-2 & \multicolumn{1}{c|}{0.8} & \multicolumn{1}{c|}{0.208} & \multicolumn{1}{c|}{0.475} & 0.116 & \multicolumn{1}{c|}{1.641} & \multicolumn{1}{c|}{0.895} & \multicolumn{1}{c|}{0.621} & 0.045 \\ \hline
\multirow{2}{*}{\begin{tabular}[c]{@{}c@{}}Amaravati, Andhra Pradesh,\\  India\end{tabular}} & Landsat & \multicolumn{1}{c|}{1.405} & \multicolumn{1}{c|}{0.351} & \multicolumn{1}{c|}{0.528} & 0.104 & \multicolumn{1}{c|}{1.598} & \multicolumn{1}{c|}{0.515} & \multicolumn{1}{c|}{0.62} & 0.031 \\ \cline{2-10} 
 & Sentinel-2 & \multicolumn{1}{c|}{1.291} & \multicolumn{1}{c|}{0.778} & \multicolumn{1}{c|}{0.518} & 0.088 & \multicolumn{1}{c|}{2.426} & \multicolumn{1}{c|}{2.821} & \multicolumn{1}{c|}{0.617} & 0.05 \\ \hline
\multirow{2}{*}{\begin{tabular}[c]{@{}c@{}}Tarlac, Luzon, \\ Phillipines\end{tabular}} & Landsat & \multicolumn{1}{c|}{0.693} & \multicolumn{1}{c|}{0.129} & \multicolumn{1}{c|}{0.361} & 0.11 & \multicolumn{1}{c|}{1.277} & \multicolumn{1}{c|}{0.246} & \multicolumn{1}{c|}{0.635} & 0.024 \\ \cline{2-10} 
 & Sentinel-2 & \multicolumn{1}{c|}{0.788} & \multicolumn{1}{c|}{0.162} & \multicolumn{1}{c|}{0.486} & 0.11 & \multicolumn{1}{c|}{1.207} & \multicolumn{1}{c|}{0.246} & \multicolumn{1}{c|}{0.622} & 0.044 \\ \hline
\multirow{2}{*}{\begin{tabular}[c]{@{}c@{}}Melbourne, Victoria,\\  Australia\end{tabular}} & Landsat & \multicolumn{1}{c|}{0.968} & \multicolumn{1}{c|}{0.309} & \multicolumn{1}{c|}{0.613} & 0.043 & \multicolumn{1}{c|}{1.076} & \multicolumn{1}{c|}{0.16} & \multicolumn{1}{c|}{0.649} & 0.031 \\ \cline{2-10} 
 & Sentinel-2 & \multicolumn{1}{c|}{0.925} & \multicolumn{1}{c|}{0.253} & \multicolumn{1}{c|}{0.542} & 0.095 & \multicolumn{1}{c|}{1.31} & \multicolumn{1}{c|}{0.144} & \multicolumn{1}{c|}{0.626} & 0.033 \\ \hline
\multirow{2}{*}{\begin{tabular}[c]{@{}c@{}}Meridian, Idaho,\\  United States\end{tabular}} & Landsat & \multicolumn{1}{c|}{0.939} & \multicolumn{1}{c|}{0.244} & \multicolumn{1}{c|}{0.521} & 0.172 & \multicolumn{1}{c|}{1.154} & \multicolumn{1}{c|}{0.724} & \multicolumn{1}{c|}{0.579} & 0.138 \\ \cline{2-10} 
 & Sentinel-2 & \multicolumn{1}{c|}{0.772} & \multicolumn{1}{c|}{0.128} & \multicolumn{1}{c|}{0.397} & 0.151 & \multicolumn{1}{c|}{1.088} & \multicolumn{1}{c|}{0.141} & \multicolumn{1}{c|}{0.596} & 0.047 \\ \hline
\end{tabular}}
\caption{Alignment results for a selection of 10 AOIs. The
  values are in RMSE in units of pixels. By average error as
  shown in the columns labeled ``Avg'', we mean the average
  of the reprojection errors over ALL the images in a
  stack. An AOI stack in SpaceNet-7 has around 24 Planet
  images, with one image per month. The table shows the
  errors for two very different approaches to the feature
  extraction and feature matching steps that are needed for
  for driving the alignment algorithm.  FAST+VGG represents
  a combination of the traditional FAST algorithm for
  feature extraction and VGG for feature matching.  On the
  other hand, SuperPoint for feature extraction and
  SuperGlue for feature matching are entirely based on
  neural networks.}
\label{table:murat_table}
\end{table*}

\section{Cross-Satellite Alignment Results}\label{sec:alignment_results}
As mentioned in \cref{sec:polynomials}, we use the
reprojection error to quantify the accuracy of alignment.
When aligning Planet and Landsat images, this error is
measured as the root mean squared error (RMSE) between the
projections of the keypoints in the Planet image into the
Landsat image and the corresponding keypoints in the latter.
The same logic would apply to the quantification of the
error in aligning Planet images and Sentinel images.

The first 8 rows of the \cref{table:murat_table} show
the average values for the before-and-after alignment errors
for 8 typical temporally-aligned Landsat-8 stacks vis-a-vis
the corresponding Planet stacks. The averages are over all
the roughly 24 images in the stacks.  Just to cite one AOI
to highlight the alignment accuracy achieve, for the case of
the Meridian, Idaho, AOI, the average RMSE went down
from 0.939 to 0.521 for the Landsat stack and 0.772 to 0.397
for Sentinel stack using FAST+VGG for feature extraction and
feature matching. Using SuperPoint+SuperGlue, the average
error decreased from 1.154 to 0.579 for Landsat and from
1.088 to 0.596 for Sentinel-2. 

\cref{fig:time_series} shows the first ten images in the
temporally and spatially aligned stacks of the Planet, the
Landsat, and the Sentinel images.

%-------------------------------------------------------------------------
\section{Accessing The Dataset}

This dataset is available at \url{https://engineering.purdue.edu/RVL/Database/murat/index.html}.  We have
organized the files in a manner similar to the organization
of the SpaceNet-7 dataset. For each AOI, the Landsat and the
Sentinel data are in the folders named as ``YYYY-MM" where
YYYY and MM represent the year and the month. The bands and
the metadata are stored in folders that follow the Landsat
and the Sentinel naming conventions.
%-------------------------------------------------------------------------
\section{Conclusion}
In this paper we present MuRA-T, a temporally and spatially
aligned multi-satellite dataset for research in change
detection algorithms.  We created our dataset by augmenting
the SpaceNet-7 dataset that consists of temporal sequences
of Planet images over 101 AOIs around the globe.  Our
dataset is based on the 60 AOIs meant for training.  Our
augmentation adds temporally aligned stacks of Landsat and
Sentinel images to the stacks of Planet images in the
SpaceNet-7 dataset.  Change detection algorithms can be
sensitive to even the slightest misalignments between the
images.  To achieve high precision in spatial alignment, we
used our highly flexible MuRA framework that makes it
relatively easy to experiment with different ways extracting
image features and constructing tie-point sets for the
purpose of alignment.  We hope that the MuRA-T dataset can
pave the way for even more research in the area of change
detection by leveraging the unique aspects of different
satellites.

%-------------------------------------------------------------------------
\section{Acknowledgment}
This research is based upon work supported in part by the
Office of the Director of National Intelligence (ODNI),
Intelligence Advanced Research Projects Activity (IARPA),
via Contract \#2021-21040700001. The views and conclusions
contained herein are those of the authors and should not be
interpreted as necessarily representing the official
policies, either expressed or implied, of ODNI, IARPA, or
the U.S. Government. The U.S. Government is authorized to
reproduce and distribute reprints for governmental purposes
notwithstanding any copyright annotation therein.

%%%%%%%%% REFERENCES
{\small
\bibliographystyle{ieee_fullname}
\bibliography{CVPR_murat_v14}
}

\end{document}